\documentclass[11pt,a4paper]{article}
\usepackage[hyperref]{naaclhlt2019}
\usepackage[utf8]{inputenc}
\usepackage{times}  
\usepackage{helvet}  
\usepackage{courier}  
\usepackage{url}  
\usepackage{graphicx}  
\usepackage{subcaption}  
\usepackage{latexsym}
\usepackage{multirow}
\usepackage{tikz,tikz-qtree}
\usepackage{amsmath,amsfonts,amssymb}
\usepackage[linesnumbered,ruled]{algorithm2e}
\usepackage{algorithmic}
\usepackage{verbatim}
\usepackage{CJK}
\usepackage{xcolor}
\usepackage{fancybox}
\usepackage{todonotes}

\newcommand{\eat}[1]{\ignorespaces}
\newcommand{\cut}[1]{}

\input{Definitions.tex}
\aclfinalcopy 

\usepackage{url}

\aclfinalcopy 

\newcounter{chenghao}

\newcounter{mw}

\newcounter{kt}

\newcounter{kl}

\definecolor{g-red}{HTML}{DB4437}
\definecolor{g-blue}{HTML}{4285F4}
\definecolor{g-green}{HTML}{0F9D58}
\definecolor{g-yellow}{HTML}{F4B400}
\definecolor{g-orange}{HTML}{FF9800}
\definecolor{g-grey}{HTML}{9E9E9E}

\title{Improving Span-based Question Answering Systems with Coarsely Labeled Data}

\author{Hao Cheng\textsuperscript{1}\thanks{~~This research was conducted when the author was at Google AI Language.} ~~ Ming-Wei Chang\textsuperscript{2} ~~ Kenton Lee\textsuperscript{2} ~~ Ankur Parikh \textsuperscript{2} ~~ Michael Collins\textsuperscript{2} ~~ Kristina Toutanova\textsuperscript{2} 
 \\
\textsuperscript{1} University of Washington \\ 
Seattle, WA, USA \\
\textsuperscript{2} Google AI Language, Seattle and New York, USA \\
{\tt chenghao@uw.edu} \\
{\tt \{mingweichang, kentonl, aparikh, mjcollins, kristout\}@google.com}
}

\date{}

\begin{document}
\maketitle
\begin{abstract}
We study approaches to improve fine-grained short answer Question Answering models by integrating coarse-grained data annotated for paragraph-level relevance and show that coarsely annotated data can bring significant performance gains. Experiments demonstrate that the standard multi-task learning approach of sharing representations is not the most effective way to leverage coarse-grained annotations. Instead, we can explicitly model the latent fine-grained short answer variables and optimize the marginal log-likelihood directly or use a newly proposed \emph{posterior distillation} learning objective. Since these latent-variable methods have explicit access to the relationship between the fine and coarse tasks, they result in significantly larger improvements from coarse supervision. 

\eat{
We study approaches for learning neural models predicting fine-grained labels, based on small datasets of finely annotated data and larger datasets of more readily available data with coarse annotations. We compare the standard multi-task learning approach to train shared representations to methods that explicitly model the relationship between the fine and coarse tasks by marginalization over the hidden fine-grained labels. We apply direct optimization of the marginal log-likelihood, batch EM, and introduce a new related algorithm we term \textit{posterior distillation}. On two tasks: document-level question answering with short and long answer supervision, and super-tagging with fine and coarse tag annotations, we show that explicitly modeling the dependencies between the fine and coarse labels brings significant improvements over the multi-tasking approach. 
}

\end{abstract}

\section{Introduction}
Question answering (QA) systems can provide most value for users by showing them a fine-grained short answer (answer span) in a context that supports  the answer (paragraph in a document).
However, fine-grained short answer annotations for question answering are costly to obtain, whereas non-expert annotators can annotate coarse-grained passages or documents faster and with higher accuracy.
In addition, coarse-grained annotations are often freely available from community forums such as Quora.\footnote{\url{https://www.quora.com/}} Therefore, methods that can learn to select short answers based on more abundant coarsely annotated paragraph-level data can potentially bring significant improvements. As an example of the two types of annotation, Figure 1 shows on the left a question with corresponding short answer annotation (underlined short answer) in a document, and on the right a question with a document annotated at the coarse-grained paragraph relevance level. 

In this work we study methods for learning short answer models from small amounts of data annotated at the short answer level and larger amounts of data annotated at the paragraph level. \newcite{min-seo-hajishirzi:2017:Short} recently studied a related problem of transferring knowledge from a fine-grained QA model to a coarse-grained model via multi-task learning and showed that finely annotated data can help improve performance on the coarse-grained task. We investigate the opposite and arguably much more challenging direction: improving fine-grained models using coarse-grained data.

\begin{figure*}[htp]
\label{fig:qafinecoarse}
\begin{center}
\includegraphics[width=\textwidth]{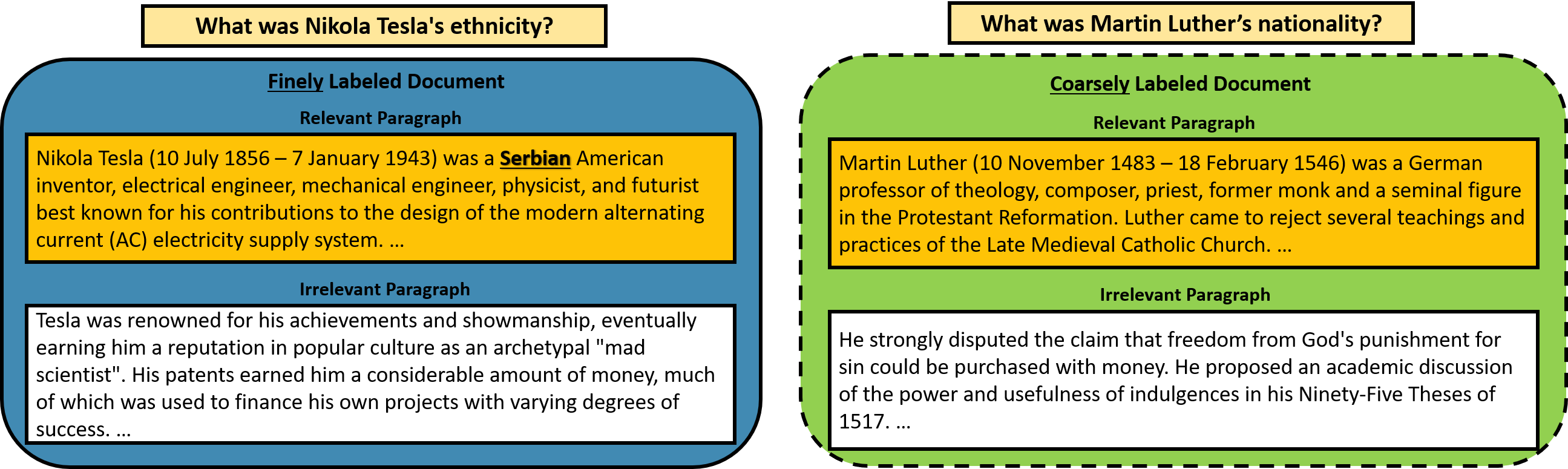}
\end{center}
\caption{An illustration of  question and answer pairs with fine-grained short answer annotation (left) and coarse-grained paragraph-level annotation (right). The finely labeled data includes both passage relevance and labeled short answer spans (\underline{Serbian} in the example), while the coarsely labeled data only provides
labels at the paragraph level.}
\end{figure*}


We explore alternatives to the standard approach of multi-task learning via representation sharing~\cite{Collobert2008ICML} by leveraging the known correspondences between the coarse and fine-grained tasks. In the standard representation sharing approach, the dependencies between the fine-grained and coarse-grained tasks are modeled {\em implicitly}. The model must learn representations that are useful for all tasks without knowing how they relate to each other. However, in the scenario of learning from both fine and coarse supervision, the dependencies between the tasks can be modeled {\em explicitly}. For example, if a paragraph answers a question, we know that there exists a fine-grained answer span in the paragraph, providing strong constraints on the possible fine-grained answers for the question. 


We evaluate a multi-task approach and three algorithms that explicitly model the task dependencies. We perform experiments on document-level variants of the SQuAD dataset~\cite{squad}. The contributions for our papers are:

\begin{itemize}
    \item We show, for the first time, that it is possible to transfer knowledge from coarsely labeled data (paragraph-level) to a fine-grained (span-based) neural QA model.\footnote{Previous work~\cite{min-seo-hajishirzi:2017:Short}  has shown that it is possible to transfer knowledge
    from finely labeled data to a coarse-grained QA task, but not the other way around.}
    The best method of using coarse-grained annotation improves performance over models using only finely labeled data by 3.8 points absolute, achieving 41\% of the improvement that could be obtained with the same amount of finely annotated data.
    \item When learning from both fine-grained and coarse-grained supervision signals, we found that latent variable models perform significantly better compared to the multi-task learning algorithm.
    \item Among the latent variable models, our newly proposed posterior distillation method outperforms direct likelihood maximization and EM due to its flexibility to generalize to multiple distance functions between model and teacher predictive distributions.
    
\end{itemize}



\section{Task Definitions}
\label{sec:background}

The fine-grained short question answering task asks to select an answer span in a document containing multiple paragraphs. In the left example in Figure 1, the short answer to the question \textit{What was Nikola Tesla's ethnicity?} is the phrase \textit{Serbian} in the first paragraph in the document. 

The coarse-grained labels indicate the relevance of document paragraphs. In the right example in Figure 1, the labels
indicate whether or not the paragraphs in a given document
contain the answers for the given question \textit{What was Martin Luther's nationality?} without specifying the answer spans. 

The goal of our paper is to design methods to learn
from both fine-grained and coarse-grained labeled data, to improve systems for fine-grained QA.

\subsection{Formal Definition}
We define the fine-grained task of interest $T_y$ as predicting outputs $y$ from a set of possible outputs ${\cal{Y}}(x)$ given inputs $x$. We say that a task $T_z$ to predict outputs $z$ given inputs $x$ is a coarse-grained counterpart of $T_y$, iff each coarse label $z$ determines a sub-set of possible labels ${\cal{Y}}(z,x) \subset {\cal{Y}}(x)$, and each fine label $y$ has a deterministically corresponding single coarse label $z$. We refer to the fine-grained and coarse-grained training data as $D_y$ and $D_z$ respectively.

For our application of document-level QA, $T_y$ is the task of selecting a short answer span from the document, and $T_z$ is the task of selecting a paragraph from the document. The input $x$ to both tasks is a question-document pair. Each document is a sequence of $M$ paragraphs, and each paragraph with index $p$ (where $1 \le p \le M$) is a sequence of $n_p$ tokens. The set of possible outputs for the  fine-grained task $T_y$ is the set of all phrases (contiguous substring spans) in all document paragraphs. The possible outputs for the coarse task $T_z$ are the paragraph indices $p$. It is clear that each paragraph output $z$ determines a subset of possible outputs $y$ (the phrases in the paragraph). 


Fine-grained annotation is provided as $y=(a_{\mathit{p}}, a_{\mathit{start}}, a_{\mathit{end}})$, where $a_{\mathit{p}}$ indicates the index of the paragraph containing the answer, and $a_{\mathit{start}}, a_{\mathit{end}}$ respectively indicate the start and end position of the short answer.

Paragraph-level supervision is provided as $z=(a_{\mathit{p}}, \_,\_)$, only indicating the paragraph index of the answer, without the start and end token indices of the answer span. The coarse labels $z$ in this case limit the set of possible labels $y$ for $x$ to: 
\begin{equation*}
    \label{eq:joint_dis}
    {\cal{Y}}(z,x) = \{(a_{\mathit{p}}, a'_{\mathit{start}}, a'_{\mathit{end}})~|~1 \leq a'_{\mathit{start}} \leq a'_{\mathit{end}} \leq n_p\}.
\end{equation*}




\paragraph{MixedQA} In the presence of the coarsely annotated $D_z$ when the task of interest is $T_y$, the research question becomes: { how can we train a model to use both $D_z$ and $D_y$ in the
most effective way?}

\section{Multi-task learning for MixedQA}
\label{sec:multi-task}
The multi-task learning approach defines models for $T_y$ and $T_z$ that share some of their parameters. The data for task $T_z$ helps improve the model for $T_y$ via these shared parameters (representations). Multi-task learning with representation sharing is widely used with auxiliary tasks from  reconstruction of unlabeled data \cite{Collobert2008ICML} to machine translation and syntactic parsing \cite{luong2015multi}, and can be used with any task $T_z$ which is potentially related to the main task of interest $T_y$.

Let
$\theta = \begin{bmatrix}
\theta_y & \theta_z & \theta_{s}
\end{bmatrix}$ be the set of parameters in the two models. $\theta_y$ denotes parameters exclusive to the fine-grained task $T_y$, $\theta_z$ denotes parameters exclusive to the coarse-grained task $T_z$, and $\theta_s$ denotes the shared parameters across the two tasks.

Then the multi-task learning objective is to minimize $L(\theta, D_y,D_z)$:
\begin{equation}
\begin{split}
& -\sum_{(x,y) \in D_{y}} \log P(y|x, \theta_s, \theta_y) \\ -~~~\alpha_z &\sum_{(x,z) \in D_{z}} \log P(z|x , \theta_s, \theta_z)
\end{split}
\end{equation}

Here $\alpha_z$ is a trade-off hyper-parameter to balance the objectives of the fine and coarse models.

We apply multi-task learning to question answering by reusing the architecture from \newcite{min-seo-hajishirzi:2017:Short} to define models for both fine-grained short answer selection $T_y$ and coarse-grained paragraph selection $T_z$. After the two models are trained, only the model for the fine-grained task $T_y$ is used at test time to make predictions for the task of interest.

The shared component with parameters $\theta_s$ maps the sequence of tokens in the document $d$ to continuous representations contextualized with respect to the question $q$ and the tokens in the paragraph $p$.  We denote these representations as 
$$\mathbf{h}(x,\theta_s) = (\mathbf{h}^1(\theta_s),\mathbf{h}^2(\theta_s),\ldots,\mathbf{h}^{M}(\theta_s)),$$ where we omit the dependence on $x$ for simplicity. Each contextualized paragraph token representation is a sequence of contextualized token representations, where $$\mathbf{h}^p(\theta_s) = {h_1}^p(\theta_s),\ldots,{{h}_{n_p}}^p(\theta_s).$$

\subsection{Fine-grained answer selection model}
\label{sec:finemodel}

The fine-grained answer selection model $P(y|x,\theta_s,\theta_y)$ uses the same hidden representations $\mathbf{h}(x,\theta_s)$ and makes predictions assuming that the start and end positions of the answer are independent, as in BiDAF \cite{bidaf}. The output parameters $\theta_y$ contain separate weights for predicting starts and ends of spans:
$\theta_y =
\begin{bmatrix}
\theta_y^{\mathit{start}} & \theta_y^{\mathit{end}}
\end{bmatrix}$

The probability of answer start $a_{\mathit{start}}$ in paragraph $a_{\mathit{p}}$ is proportional to $\exp(h(a_{\mathit{start}}, a_{\mathit{p}}, \theta_s)\cdot \theta_y^{\mathit{start}})$, where $h(a_{\mathit{start}}, a_{\mathit{p}}, \theta_s)$ is the hidden representation of the token $a_{\mathit{start}}$ in paragraph $a_{\mathit{p}}$, given shared parameters $\theta_s$.
The probability for end of answer positions is defined analogously.


\subsection{Paragraph answer selection model}

The paragraph selection model for task $T_{z}$ uses the same hidden representations $\mathbf{h}(x,\theta_s)$ for the tokens in the document. Because this model assigns scores at the paragraph granularity (as opposed to token granularity), we apply a pooling operation to the token representations to derive single vector paragraph representations. As in \cite{min-seo-hajishirzi:2017:Short}, we use max-pooling over token representations and arrive at \[h^p(\theta_s)=\mbox{max}({h_1}^p(\theta_s),\ldots,{{h}_{n_p}}^p(\theta_s))\]

Using the coarse-grained task-specific parameters $\theta_z$, we define the probability distribution over paragraphs as: \[P(a_p = p | x, \theta_s, \theta_z) = \frac{\exp(h^p(\theta_s) \cdot \theta_z)}{\sum_{p'}{\exp(h^{p'}(\theta_s) \cdot \theta_z)}} \]

\section{Latent Variable Methods for MixedQA}
\begin{figure*}
    \centering
    \begin{subfigure}[b]{0.3\textwidth}
        \includegraphics[width=\textwidth, height=5cm]{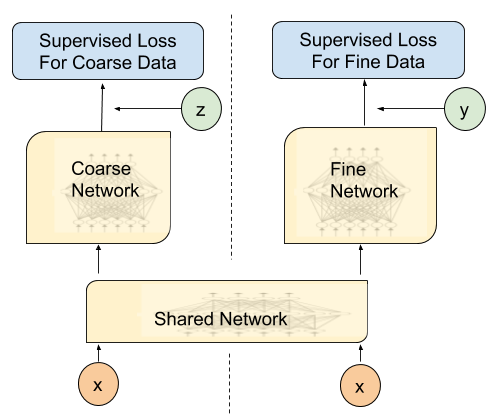}
        \caption{}
        \label{fig:mtl}
    \end{subfigure}
\hspace{1.2cm}    
    \begin{subfigure}[b]{0.6\textwidth}
        \includegraphics[width=\textwidth, height=5cm]{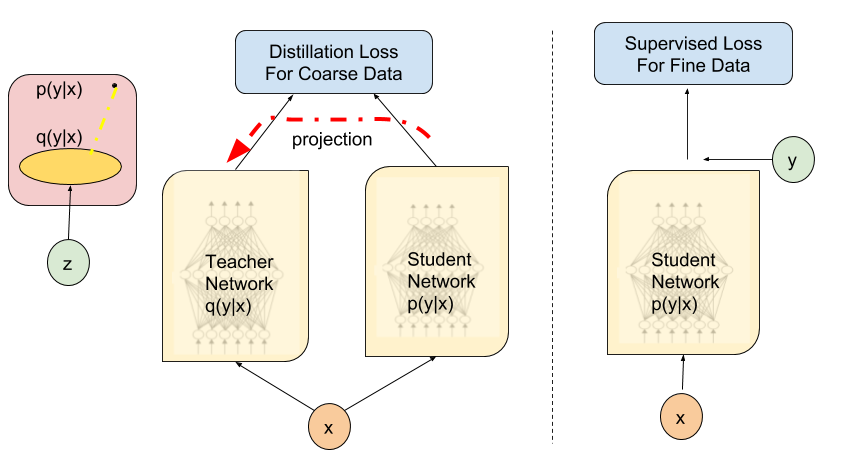}
        \caption{}
        \label{fig:dns}
    \end{subfigure}
    \caption{Illustration of the multi-task learning algorithm (a) and the posterior distillation latent variable methods (b). See text for more details. }
\end{figure*}

We study two types of latent variable methods that capture the dependencies between the fine and coarse tasks explicitly. Unlike the multitask learning algorithm described above, both eliminate the need for parameters specifically for the coarse task $\theta_z$, since we treat the fine labels as a latent variable in the coarsely annotated data.

The dependencies between the coarse and fine supervision labels can be captured by the
following consistency constraints implied by our task definition:\
\begin{equation*}
 \begin{split}
 P(y, z|x) = 0, & \forall y \notin\mathcal{Y}(z, x), \text{ and } \\
 P(z|y,x) = 1, & \forall y \in\mathcal{Y}(z, x).
 \end{split}
\end{equation*}

\subsection{Maximum Marginal Likelihood}

For the task of document-level QA, these constraints ensure that a paragraph is labeled as positive iff there exists a positive answer text span inside the paragraph. 


The idea of the maximum marginal likelihood method is to define a distribution over coarse labels using the fine-grained model's distribution over fine labels.
By expanding the above equations expressing the task dependencies,
\begin{equation}
P(z|x, \theta) = \sum_{y \in\mathcal{Y}(x)} P(y,z|x, \theta) = \hspace{-6pt}\sum_{y \in\mathcal{Y}(z, x)}\hspace{-6pt}P(y|x, \theta)
\label{eq:latentmodel}
\end{equation}

This equation simply says that the probability that a given paragraph $z$ is relevant is the sum of the probabilities of all possible short answer spans within the paragraph.

The objective function for the coarsely labeled data $D_{z}$ can be expressed as a function of the parameters of the fine-grained task model as:
\begin{equation}
\begin{split}
-\sum_{(x,z) \in D_{z}} \log \sum_{y \in\mathcal{Y}(z, x)} P(y|x, \theta_s, \theta_y)
\end{split}
\end{equation}

The fine-grained task loss and the coarse-grained task loss are interpolated with a parameter $\alpha_z$, as for the multi-task approach.

\subsection{Posterior Distillation}

In addition to direct maximization of the marginal likelihood for latent variable models \cite{salakhutdinov2003optimization}, prior work has explored EM-based optimization \cite{moon1996expectation} including generalized EM \cite{wu1983convergence}, which is applicable to neural models \cite{greff2017neural}.

We present a class of optimization algorithms which we term Posterior Distillation, which includes generalized EM for our problem as a special case, and has close connections to knowledge distillation~\cite{ba2014deep,hinton2015distilling}.

We begin by describing an online generalized EM optimization algorithm for the latent variable model from equation (\ref{eq:latentmodel}) and show how it can be generalized to multiple variants inspired by knowledge distillation with priviledged information \cite{LopezPaz2015UnifyingDA}. We refer to the more general approach as Posterior Distillation.

\begin{algorithm}[t] 
\caption{Posterior Distillation Algorithm.} 
\label{algo:PD} 
\begin{algorithmic}[1] 
    \WHILE{not converge}
        \STATE Sample a mini-batch $(x_1,y) \sim D_y$ and $(x_2,z) \sim D_z$
        \STATE Calculate predicted distribution for current $\theta^{old}$ $P(\hat{y}|x_2, \theta^{old})$
        \STATE Correct and renormalize the predicted distribution using the coarse supervision signal by setting
               $$q(\hat{y}|x_2) \propto \begin{cases}
P(\hat{y}|x_2, \theta^{old}), \hat{y} \in \mathcal{Y}(z)\\
0, \hat{y} \notin \mathcal{Y}(z)
\end{cases}$$
        \STATE Update $\theta$ by taking a step to minimize -$\log P(y|x_1, \theta)$ + $\alpha_z$\textsc{distance}($P(y|x,\theta), q$).
    \ENDWHILE
\end{algorithmic}
\end{algorithm}





In EM-like algorithms one uses current model parameters $\theta^{old}$ to make predictions and complete the latent variables in input examples, and then updates the model parameters to maximize the log-likelihood of the completed data. We formalize this procedure for our case below.

Given a coarse example with input $x$ and coarse label $z$, we first compute the posterior distribution over the fine labels $y$ given $z$ and the current set of parameters $\theta^{old}$:
\begin{align}
P(y | x, z, \theta^{old}) &= \frac{[[\hbox{$y \in {\cal Y}(x)$}]] \times P(y | x, \theta^{old})}{\displaystyle\sum_{y \in {\mathcal Y}(z, x)} P(y | x, \theta^{old})}
\label{eq:project}
\end{align}
where $[[\cdot]]$ is the indicator function.
In EM, we update the parameters $\theta$ to minimize the negative expected log-likelihood of the fine labels with respect to the posterior distribution:
\begin{align*}
Q(\theta, \theta^{old}) &= -\mathop{\mathbb{E}}_{P(y | x, z, \theta^{old})} \log P(y | x, \theta)\\
&= -\sum_{y \in {\cal Y}(x)} P(y | x, z, \theta^{old}) \log P(y | x, \theta)
\end{align*}
By taking a gradient step towards minimizing $Q(\theta, \theta^{old})$ with respect to $\theta$, we arrive at a form of generalized EM~\cite{wu1983convergence}. If the loss $Q$ is computed over a mini-batch, this is a form of online EM.

We propose a variant of this EM algorithm that is inspired by knowledge distillation methods~\cite{ba2014deep,hinton2015distilling}, where a student model learns to minimize the distance between its predictions and a teacher model's predictions. In our case, we can consider the posterior distribution $P(y | x, z, \theta^{old})$ to be the teacher, and the model distribution $P(y | x, \theta)$ to be the student. Here the teacher distribution is directly derived from the model (student) distribution $P(y | x, \theta^{old})$ by integrating the information from the coarse label $z$. The coarse labels can be seen as privileged information \cite{LopezPaz2015UnifyingDA} which the student does not condition on directly.

Let us define $Q(\theta, \theta^{old})$ in a more general form, where it is a general distance function rather than cross-entropy:
\[
Q(\theta, \theta^{old}) = \textsc{distance}(P(y | x, z, \theta^{old}), P(y | x, \theta))
\]
 We refer to the class of learning objectives in this form as \emph{posterior distillation}. When the distance function is cross entropy, posterior distillation is equivalent to EM. As is common in distillation techniques~\cite{NIPS2014_5484}, we can apply other distance functions, such as the squared error.
\[
Q(\theta, \theta^{old}) = \sum_{y \in {\cal Y}(x)} \left\lVert P(y | x, z, \theta^{old}) - P(y | x, \theta) \right\rVert_2^2
\]
In our experiments, we found that squared error outperforms cross entropy consistently.

This algorithm also has a close connection to Posterior Regularization \cite{ganchev2010posterior}. The coarse supervision labels $z$ can be integrated using linear expectation constraints on the model posteriors $P(y|x,\theta)$, and a KL-projection onto the constrained space can be done exactly in closed form using equation \ref{eq:project}. Thus the PR approach in this case is equivalent to posterior distillation with cross-entropy and to EM. Note that the posterior distillation method is more general because it allows additional distance functions.

The combined loss function using both finely and coarsely labeled data to be minimized is:

\begin{equation}
\begin{split}
& \sum_{(x,y) \in D_{y}} -\log P(y|x, \theta_s) \\ +~~~\alpha_z &\sum_{(x,z) \in D_{z}} Q(\theta,\theta^{old},x,z) 
\end{split}
\end{equation}

Figure 2 presents an illustration of the multi-task and posterior distillation approaches for learning from both finely and coarsely labeled data. Algorithm 1 lists the steps of optimization. Each iteration of the loop samples mini-batches from the union of  finely and coarsely labeled data and takes a step to minimize the combined loss.

\section{Experiments}

We present experiments on question answering using the multi-task and latent variable methods introduced in the prior section.



\subsection{Mixed supervision data}
We focus on the document-level variant of the SQuAD dataset~\cite{squad}, as defined by \newcite{docqa}, where given a question and document, the task is to determine the relevant passage and answer span within the passage $(a_p, a_{\mathit{start}}, a_{\mathit{end}})$. We define finely annotated subsets $D_{y}$ with two different sizes: 5\% and 20\% of the original dataset. These are paired with non-overlapping subsets of coarsely annotated data $D_{z}$ with sizes 20\% and 70\% of the original training set, respectively. Both of these settings represent the regime where coarsely annotated data is available in higher volume, because such data can be obtained faster and at lower cost.  For both dataset settings, we derive $D_{y}$ and $D_{z}$ from the SQuAD training set, by allocating whole documents with all their corresponding questions to a given subset. In both settings, we also reserve a finely annotated non-overlapping set $\mbox{Dev}_{y}$, which is used to select optimal hyperparameters for each method.\footnote{We reserve 10\% of the data for $\mbox{Dev}_{y}$, and thus we only train with up to 90\% of the SQuAD training set.} We report final performance metrics on $\mbox{Test}_{y}$, which is the unseen SQuAD development set.


\subsection{QA model}
We build on the state-of-the-art publicly available question answering system by \newcite{docqa}.\footnote{\url{https://github.com/allenai/document-qa}} The system extends BiDAF \cite{bidaf} with self-attention
and performs well on document-level QA. We reuse all hyperparameters from \newcite{docqa} with the exception of number of paragraphs sampled in training: 8 instead of 4. Using more negative examples was important when learning from both fine and coarse annotations. The model uses character embeddings with dimension 50, pre-trained Glove embeddings, and hidden units for bi-directional GRU encoders with size 100. Adadelta is used for optimization for all methods.
We tune two hyperparameters separately for each condition based on the held-out set: (1) $\alpha \in \{.01, .1, .5, 1, 5, 10, 100 \}$, the weight of the coarse loss, and (2) the number of steps for early stopping. The training time for all methods using both coarse and fine supervision is comparable.  We use Adadelta for optimization for all methods.
\subsection{Results}

We report results evaluating the impact of using coarsely annotated data in the two dataset conditions in Figure \ref{tab:qaresultsfinal}. There are two groups of rows corresponding to the two data sizes: in the smaller setting, only 5\% of the original fine-grained data is used, and in the medium setting, 20\% of the fine-grained data is used.   The first row in each group indicates the performance when using only finely labeled fully supervised data. The column Fine-F1 indicates the performance metric of interest -- the test set performance on document-level short answer selection. The next rows indicate the performance of a multi-task and the best latent variable method when using the finely labeled data plus the additional coarsely annotated datasets. The ceiling performance in each group shows the oracle achieved by a model also looking at the gold fine-grained labels for the data that the rest of the models see with only coarse paragraph-level annotation. The column \textbf{Gain} indicates the relative error reduction of each model compared to the supervised-only baseline with respect to the ceiling upper bound. As we can see all models benefit from coarsely labeled data and achieve at least 20\% error reduction. The best latent variable method (Posterior Distillation with squared error distance) significantly outperforms the multi-task approach, achieving up to 41\% relative gain.



Figure \ref{tab:qaresults} compares the performance of the three different optimization methods using latent fine-grained answer variables for coarsely annotated data. Here we inlcude an additional last column reporting performance on an easier task where the correct answer paragraph is given at test time, and the model only needs to pick out the short answer within the given paragraph. We include this measurement to observe whether models are improving just by picking out relevant paragraphs or also by selecting the finer-grained short answers within them.
Since EM and MML are known to optimize the same function, it is unsurprising that MML and PD with cross-entropy (equivalent to EM) perform similarly. For posterior distillation, we observe substantially better performance with the squared error as the distance function, particularly in the second setting, where there is more coarsely annotated data.

\begin{figure*}[ht]
    \centering
    \begin{tabular}{|l|l|c|r|}
    \hline
         \textbf{Data} & \textbf{Model} & \textbf{{Fine-F1}} & \textbf{Gain} \\
         \hline
          D$_{5fine}$ & Supervised & 50.3 & 0.0\% \\
          D$_{5fine}$ + D$_{20coarse}$ & MTL & 53.2 & 21.0\% \\
         D$_{5fine}$ + D$_{20coarse}$ & PD (err$^2$) & \textbf{54.9} & \textbf{33.3}\% \\
          \hline
          D$_{25fine}$ & Ceiling& \text{64.1} & 100.0\% \\
         \hline\hline
          D$_{20fine}$ & Supervised & 62.0 & 0\%  \\
          D$_{20fine}$ + D$_{70coarse}$ & MTL & 64.2 & 23.9\%  \\
          D$_{20fine}$ + D$_{70coarse}$ & PD (err$^2$) & \textbf{65.8} & \textbf{41.3\%}  \\
          \hline
          D$_{90fine}$ & Ceiling& \text{71.2} & 100.0\%\\
         \hline
    \end{tabular}
    \caption{Results on short answer selection at the document level comparing the performance of models using fine-only data to ones also using coarsely labeled data. Contrasting multi-task to the best method using latent fine answer variables.
    The relative gains over the fine only baseline with. respect to the ceiling are shown in the "Gain" column.}
    \label{tab:qaresultsfinal}
\end{figure*}

\eat{
\begin{table}[ht]
    \centering
    \begin{tabular}{|l|l|c|}
    \hline
         \textbf{Data} & \textbf{Model} & \textbf{{Doc-F1}} \\
         \hline
       
          D$_{5f}$ & Fine only & 50.3 (0.0\%) \\
         D$_{5f}$ + D$_{20c}$ & MTL & 53.2 (21.0\%) \\
        
          D$_{5f}$ + D$_{20c}$ & PD (err$^2$) & \textbf{54.9} (33.3\%) \\
          \hline
          D$_{25f}$ & Ceiling& \textit{64.1} (100\%) \\
         \hline\hline
          D$_{20f}$ & Fine only & 62.0 (0\%)  \\
          
        D$_{20f}$ + D$_{70c}$ & MTL & 64.2 (23.9\%)  \\
        
          D$_{20f}$ + D$_{70c}$ & PD (err$^2$) & \textbf{65.8} (41.3\%)  \\
          \hline
          D$_{90f}$ & Ceiling& \textit{71.2} (100\%)\\
         \hline
    \end{tabular}
    \caption{Results on short answer selection at the document level comparing the performance of models using fine-only data to ones also using coarsely labeled data. Contrasting multi-task to the best method using latent fine answer variables. The standard deviation via five different random initializations is indicated.}
    \label{tab:qaresultsfinal}
\end{table}
}

To gain more insight into the behavior of the different methods using coarsely annotated data, we measured properties of the predictive distributions $P(y|x,\theta)$ for the three methods on the dataset used with coarse labels in training $D_{70coarse}$. The results are shown in Figure \ref{tab:analysis}. For models MTL, MML, PD($xent$), and PD($err^2$), trained on finely labeled $D_{20fine}$ and coarsely labeled $D_{70coarse}$, we study the predictive distributions $P(y|x,\theta^M)$ for the four model types $M$. We measure the properties of these distributions on the dataset $D_{70fine}$, which is the finely labeled version of the same (question, document)-pairs $D_{70}$ as $D_{70coarse}$. Note that none of the models see the fine-grained short answer labels for $D_{70}$ in training since they only observe paragraph-level relevance annotations. Nevertheless, the models can assign a probability distribution over fine-grained labels in the documents, and we can measure the peakiness (entropy) of this distribution, as well as see how it compares to the gold hidden label distribution.

The first column in the table reports the entropies of the predictive distributions for the four trained models (using the fine task model for the multi-task method MTL). We can see that multi-task method MTL and PD($xent$) (which is equivalent to generalized EM) have lowest entropy, and are most confident about their short answer predictions. MML marginalizes over possible fine answers, resulting in flatter predictive distributions which spread mass among multiple plausible answer positions. The best-performing method PD($err^2$) is somewhere in between and maintains more uncertainty. The next two columns in the Table look at the cross-entropy ($xent$) and squared error  ($err^2$) distances of the predictive distributions with respect to the gold one. The gold label distribution has mass of one on a single point indicating the correct fine answer positions. Note that none of the models have seen this gold distribution during training and have thus not been trained to minimize these distances (the PD latent variable models are trained to minimize distance with respect to projected model distributions given coarse passage labels $z$).  We can see that the predictive distribution of the best method PD($err^2$) is closest to the gold labels. The maximum marginal likelihood method MML comes second in approaching the gold distribution. The multi-task approach lags behind others in distance to the fine-grained gold labels, but comes first in the measurement in the last column, Passage-MRR. That column indicates the mean reciprocal rank of the correct gold {\em{passage}} according to the model. Here passages are ranked by the score of the highest-scoring short answer span within the passage. This measurement indicates that the multi-task model is able to learn to rank passages correctly from the coarse-grained passage-level annotation, but has a harder time to transfer this improvement to the task of picking fine-grained short answers within the passages.

\begin{figure*}[ht]
    \centering
    \begin{tabular}{|l|l|c|c|}
    \hline
         \textbf{Data} & \textbf{Model} & \textbf{{Fine-F1}} & \textbf{Fine Passage-F1} \\
         \hline
      
         D$_{5fine}$ + D$_{20coarse}$ & MML & 54.3 ($\pm$ 0.7) & 62.0 ($\pm$ 1.1)  \\
         D$_{5fine}$ + D$_{20coarse}$ & PD (xent) & 54.2 ($\pm$ 0.5) & 62.3 ($\pm$ 0.8)  \\
          D$_{5fine}$ + D$_{20coarse}$ & PD (err$^2$) & \textbf{54.9} ($\pm$ 0.6)  & \textbf{63.0} ($\pm$ 0.6) \\ 
         \hline\hline
         D$_{20fine}$ + D$_{70coarse}$ & MML & 64.9 ($\pm$ 0.2) & 72.4 ($\pm$ 0.4)   \\
          D$_{20fine}$ + D$_{70coarse}$ & PD (xent) & 64.8 ($\pm$ 0.2) & 72.5 ($\pm$ 0.2)   \\
          D$_{20fine}$ + D$_{70coarse}$ & PD (err$^2$) & \textbf{65.8} ($\pm$ 0.3) &  \textbf{73.1} ($\pm$ 0.3) \\
         \hline
    \end{tabular}
    \caption{
    Comparison between different latent variable methods. We report the standard deviation via five different random initialization. Note that PD($err^2$) is also the best algorithm when the passage is given.
    }
    \label{tab:qaresults}
\end{figure*}

\eat{
\begin{table}[ht]
    \centering
    \begin{tabular}{|l|l|c|c|}
    \hline
         \textbf{Data} & \textbf{Model} & \textbf{{Doc-F1}} \\
         \hline
         \eat{
        D$_{10f}$ & Fine only & 56.5 \\ 
        D$_{10f}$ + D$_{40c}$ & MTL & 57.2 \\ 
         D$_{10f}$ + D$_{40c}$ & Marginalize & 61.0 \\
          D$_{10f}$ + D$_{40c}$ & EM & 61.1 \\
          D$_{10f}$ + D$_{40c}$ & Posterior Dist. & 61.6 \\
          D$_{50f}$ & Fine only ceiling & & \\
          \hline
          }
        
          D$_{5f}$ & Fine only & 50.3 ($\pm$ 0.4) & 59.7 ($\pm$ 0.4) \\
         D$_{5f}$ + D$_{20c}$ & MTL & 53.2 ($\pm$ 0.5) & 62.0 ($\pm$ 0.5)  \\ 
        \hline
         D$_{5f}$ + D$_{20c}$ & MML & 54.3 ($\pm$ 0.7) & 62.0 ($\pm$ 1.1)  \\
         D$_{5f}$ + D$_{20c}$ & PD (xent) & 54.2 ($\pm$ 0.5) & 62.3 ($\pm$ 0.8)  \\
          D$_{5f}$ + D$_{20c}$ & PD (err$^2$) & \textbf{54.9} ($\pm$ 0.6)  & \textbf{63.0} ($\pm$ 0.6) \\ \hline
          D$_{25f}$ & Ceiling& \textit{64.1} ($\pm$ \textit{0.4}) & \textit{73.3} ($\pm$ \textit{0.3})\\
         \hline\hline
          D$_{20f}$ & Fine only & 62.0 ($\pm$ 0.7)  & 71.2 ($\pm$ 0.5)    \\
        D$_{20f}$ + D$_{70c}$ & MTL & 64.2 ($\pm$ 0.3)  &  71.9 ($\pm$ 0.3)  \\ 
        \hline
         D$_{20f}$ + D$_{70c}$ & MML & 64.9 ($\pm$ 0.2) & 72.4 ($\pm$ 0.4)   \\
          D$_{20f}$ + D$_{70c}$ & PD (xent) & 64.8 ($\pm$ 0.2) & 72.5 ($\pm$ 0.2)   \\
          D$_{20f}$ + D$_{70c}$ & PD (err$^2$) & \textbf{65.8} ($\pm$ 0.3)   &  \textbf{73.1} ($\pm$ 0.3) \\ \hline
          D$_{90f}$ & Ceiling& \textit{71.2} (100\%)\\
         \hline
    \end{tabular}
    \caption{Results on short answer selection at the document and passage level using different combinations of fine and coarse supervision. We also report the standard deviation via five different random initialization.}
    \label{tab:qaresults}
\end{table}

}

\begin{figure*}[htp!]
    \centering
    \begin{tabular}{|l|c|c|c|c|}
    \hline
     \textbf{Model} & \textbf{Entropy} & \textbf{$xent$-Gold} & \textbf{$err^2$-Gold} & \textbf{Passage-MRR} \\ \hline
     MTL &  1.53 & 2.01 & .630 &   \textbf{94.3} \\
     MML &  1.89 & 1.89 & .605  & 93.6 \\
     PD($xent$) & 1.59 & 2.09 & .635  & 92.2 \\
     PD($err^2$) & 1.68 & \textbf{1.87} & \textbf{.592} & 94.1 \\
     \hline
     \end{tabular}
     \caption{Properties of predictive distributions of coarsely annotated data
			 for different models trained on  finely labeled $D_{20f}$ and coarsely
			 labeled $D_{70c}$. The measurements are on the finely labeled version of
			 $D_{70c}$. \textbf{Entropy} measures how uncertain the models are about
			 the short answer locations. \textbf{$xent$-Gold} and
			 \textbf{$err^2$-Gold} measure the cross-entropy and squared error
			 distance to the gold fine-grained label distribution. \textbf{Passage-MRR} measures the ability of models to identify the relevant paragraph.
     }
     \label{tab:analysis}
     \end{figure*}




\eat{
 
 }

\section{Related Work}


\subsection{Text-based Question Answering}
In span-based reading comprehension, a system must be able to extract a
plausible text-span answer for a given question from a context document or
paragraph \cite{squad,joshi-EtAl:2017:Long,Trischler2016arxiv}.  Most work has focused on selecting short answers given relevant paragraphs, but datasets and works considering the more realistic task of selection from full documents are starting to appear \cite{joshi-EtAl:2017:Long}.


Sentence selection or paragraph selection datasets test whether a system can correctly rank texts
that are relevant for answering a question higher than texts that do not.
\newcite{Wang2007EMNLP} constructed the QASent dataset based on questions from
TREC 8-13 QA tracks. WikiQA \cite{yang-yih-meek:2015:EMNLP} associates questions
from Bing search query log with all the sentences in the Wikipedia summary
paragraph which is then labeled by crowd workers.  Most state-of-the-art models for both types of tasks make use of neural network modules to construct and compare representations for a question and the possible answers. We build on a near state-of-the-art baseline model and evaluate on a  document-level short question answering task.

\subsection{Data Augmentation and Multi-Task Learning in QA} 
There have been several works addressing the paucity of annotated data for QA.
Data noisily annotated with short answer spans has been generated automatically through
distant supervision and shown to be useful  \cite{joshi-EtAl:2017:Long}.
Unlabeled text and data augmentation through machine translation have been used
to improve model quality~\cite{yang-EtAl:2017:Long,peters2018deep,yu2018qanet}.
\newcite{min-seo-hajishirzi:2017:Short} used short-answer annotations in SQuAD
\cite{squad} to improve paragraph-level question answering for
WikiQA~\cite{yang-yih-meek:2015:EMNLP}. To the best of our knowledge, there has
been no prior work using QA data annotated at the paragraph level to improve
models for short question answering. 


\subsection{Learning from Weak Annotation}

Modeling task dependencies has been researched under two related frameworks, 
{\it multi-task learning} \cite{caruana1998multitask} and {\it multiple-instance learning}
\cite{maron1998framework}. Multi-task learning models have been shown to benefit
from data regularities through the implicit modeling of task dependencies, such
as representation sharing \cite{Collobert2008ICML} and parameter regularization
\cite{Duong2015ACL-IJCNLP}. Recent works have started to design more structured representation sharing based on linguistic
hierarchies \cite{Sogaard2016ACL,Hashimoto2017emnlp}. In contrast, works on
multiple-instance learning focus on  explicit reasoning over possible
fine-grained annotations for coarsely labeled examples and have been successfully applied to problem with weakly or
coarsely annotated data, such as entity and relation extraction from distant
supervision
\cite{tsuboi-EtAl:2008:PAPERS,Surdeanu:2012:MML:2390948.2391003,zeng-EtAl:2015:EMNLP}. 


Neither framework has been studied for learning from
span and paragraph annotation for short answer QA and the two frameworks have not been directly compared before. 




\section{Conclusion}

In this paper we showed that data annotated at the coarse-grained paragraph relevance level can be used to  improve the performance of a fine-grained short answer QA system, achieving 41\% of the gain that could be obtained with an equivalent amount of finely annotated data. We presented the first experimental comparison of multi-task and latent variable models for using coarsely annotated data for QA and showed that the latent variable models, which explicitly model the relationship between the fine-grained and coarse-grained relevance tasks outperform the multi-tasking approach. Finally, we showed that a distillation formulation naturally leads to considering  loss functions other than cross-entropy, resulting in significantly improved performance for distillation with a squared error loss.

In the future, we plan to study active learning algorithms to select from fine-grained and coarse-grained examples to annotate, to minimize the annotation cost. We would also like to examine the effectiveness of using large-scale coarsely labeled datasets such as the Quora community website to improve fine-grained QA models. In addition, we plan to measure the annotation cost and agreement for fine and coarse annotations.


\bibliographystyle{acl_natbib}

\end{document}